\documentclass{article}
\usepackage[nohyperref,accepted]{icml2024}

\usepackage{microtype}
\usepackage{graphicx}
\usepackage{booktabs}
\usepackage{subcaption}
\usepackage{hyperref}
\usepackage{lipsum}

\usepackage{amsmath}
\usepackage{amssymb}
\usepackage{mathtools}
\usepackage{amsthm}

\usepackage[capitalize,noabbrev]{cleveref}

\theoremstyle{plain}
\newtheorem{theorem}{Theorem}[section]

\newtheorem{lemma}[theorem]{Lemma}

\newtheorem{observation}[theorem]{Observation}
\theoremstyle{definition}

\theoremstyle{remark}

\usepackage[textsize=tiny]{todonotes}

\newcommand{\name}{\textsc{Flora}\xspace}
\newcommand{\longname}{\textbf{f}rom \textbf{L}oRA t\textbf{o} high-\textbf{r}ank upd\textbf{a}tes}
\newcommand{\mytitle}{\name: Low-Rank Adapters Are Secretly Gradient Compressors}
\icmltitlerunning{\mytitle}
\usepackage{amsmath,amsfonts,amsthm,bm,xspace,multirow}
\definecolor{myred}{RGB}{255,127,127}
\definecolor{mygreen}{RGB}{127,255,127}

\usepackage{thmtools}
\usepackage{thm-restate}

\def\1{\bm{1}}

\DeclareMathAlphabet{\mathsfit}{\encodingdefault}{\sfdefault}{m}{sl}
\SetMathAlphabet{\mathsfit}{bold}{\encodingdefault}{\sfdefault}{bx}{n}

\def\gL{{\mathcal{L}}}

\def\gN{{\mathcal{N}}}

\def\sR{{\mathbb{R}}}

\def\sW{{\mathbb{W}}}

\def\sZ{{\mathbb{Z}}}

\newcommand{\E}{\mathop{\mathbb{E}}}

 \renewcommand{\algorithmiccomment}[1]{$\rhd$ #1}

\DeclareRobustCommand{\logo}{%
  \begingroup\normalfont
  \raisebox{-0.25em}{%
  \hspace{-0.5em}
  \includegraphics[height=1.5em]{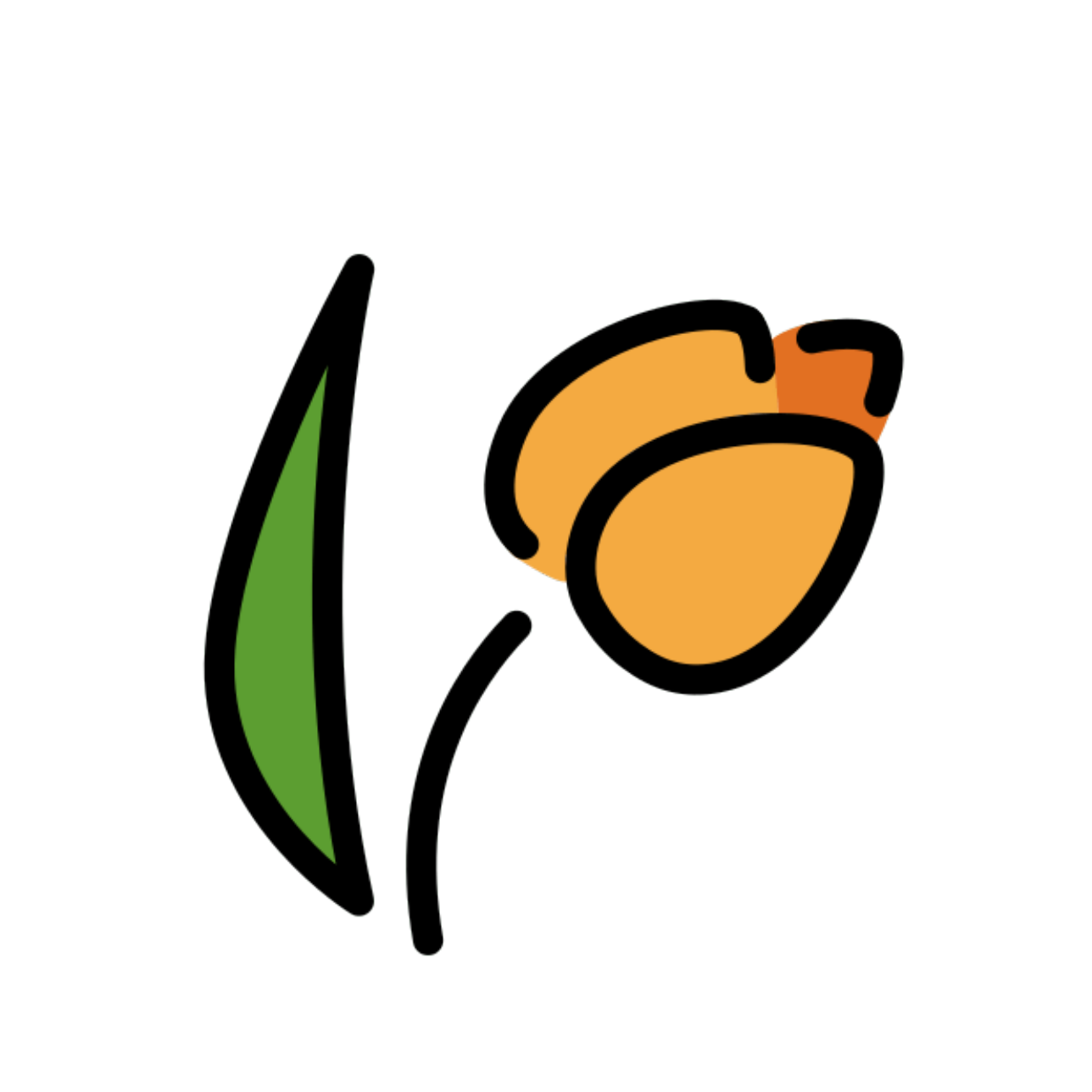}%
  }%
  \kern 0.2em
  \endgroup
}

\begin{document}

\twocolumn[
\icmltitle{\logo \mytitle}
\icmlsetsymbol{equal}{*}

\begin{icmlauthorlist}
\icmlauthor{Yongchang Hao$^*$}{uofa}
\icmlauthor{Yanshuai Cao}{bai}
\icmlauthor{Lili Mou}{uofa,cifar}
\end{icmlauthorlist}

\icmlaffiliation{uofa}{Department of Computing Science \& Alberta Machine Intelligence Institute (Amii), University of Alberta}
\icmlaffiliation{bai}{Borealis~AI}
\icmlaffiliation{cifar}{Canada CIFAR AI Chair}

\icmlcorrespondingauthor{Yongchang Hao}{yongcha1@ualberta.ca}
\icmlcorrespondingauthor{Yanshuai Cao}{yanshuai.cao@borealisai.com}
\icmlcorrespondingauthor{Lili Mou}{double power.mou@gmail.com}

\icmlkeywords{}

\vskip 0.3in
]

\printAffiliationsAndNotice{$^*$Project done during Mitacs internship at Borealis~AI.}
\begin{abstract}
Despite large neural networks demonstrating remarkable abilities to complete different tasks, they require excessive memory usage to store the optimization states for training. To alleviate this, the low-rank adaptation (LoRA) is proposed to reduce the optimization states by training fewer parameters. However, LoRA restricts overall weight update matrices to be low-rank, limiting the model performance. In this work, we investigate the dynamics of LoRA and identify that it can be approximated by a random projection. Based on this observation, we propose \name, which is able to achieve high-rank updates by resampling the projection matrices while enjoying the sublinear space complexity of optimization states. We conduct experiments across different tasks and model architectures to verify the effectiveness of our approach.
\end{abstract}

\section{Introduction}\label{sec:intro}

Gradient-based optimization powers the learning part of deep neural networks. In the simplest form, stochastic gradient descent (SGD) updates the model parameters using noisy estimation of the negative gradient. More advanced methods track various gradient statistics to stabilize and accelerate training~\cite{duchi2011adaptive, hinton2012neural}. For example, the momentum technique tracks an exponential moving average of gradients for variance reduction~\cite{cutkosky2019momentum} and damping~\cite{goh2017why}. On the other hand, gradient accumulation computes the average of gradients in the last few batches to simulate a larger effective batch for variance reduction~\cite{wang2013variance}. Both cases require an additional memory buffer equal to the model size to store the information.

However, such a linear space complexity of optimization states becomes problematic in modern deep learning. For example, GPT-3~\cite{brown2020language} and Stable Diffusion~\cite{rombach2022high} are trained with Adam~\cite{kingma2014adam} where momentum is applied. For each scalar in the parameter set, Adam maintains two additional variables (i.e., first- and second-moment estimates), tripling the memory usage. The largest GPT-3, for example, has $175$ billion parameters taking $700$GB of memory. Adam requires an additional $1.4$TB memory for optimization states. This excessive amount of memory usage poses a scaling challenge.

One line of research saves memory by training a subset of parameters~\citep{houlsby2019parameterefficient, zaken2022bitfit}, so the optimizer only stores information about a small set of trainable parameters. One notable example is the low-rank adaptation (LoRA, \citeauthor{hu2022lora}, \citeyear{hu2022lora}). LoRA updates parameter matrices by low-rank patches, which contain much fewer trainable parameters. In this way, the momentum and gradient accumulation also have much smaller sizes. However, LoRA restricts the weight update to be in the low-rank form, limiting the optimization space of the model parameters.

Another line of work designs new optimizers that use less memory~\cite{dettmers20218bit, feinberg2023sketchy}. For instance, Adafactor~\cite{shazeer2018adafactor} leverages the closed-form solution of generalized Kullback--Leibler divergence~\cite{finesso2006nonnegative} to reconstruct the second-moment estimate in Adam. To optimize a matrix in $\mathbb R^{n \times m}$, Adafactor reduces the memory from $O(nm)$ to $O(n + m)$, making the space complexity of second-moment estimation sublinear in model size. However, Adafactor drops the momentum technique to achieve the sublinearity, sacrificing the variance reduction and damping effect of momentum~\cite{rae2021scaling}. Moreover, it does not reduce the memory for gradient accumulation.

In this work, we propose \name (\longname), which is a novel optimization technique that uses sublinear memory for gradient accumulation and momentum calculation. Our intuition arises from investigating LoRA and observing that a LoRA update is dominated by a random projection, which compresses the gradient into a lower-dimensional space~\cite{dasgupta2000experiments, bingham2001random}. Thus, we propose \name that applies such a compression technique directly to the update of the original weight matrix. Our \name resamples the random projection and is able to mitigate the low-rank limitation of LoRA. Further, our approach only stores the compressed gradient accumulation and momentum, thus saving the memory usage of optimization states to the sublinear level.  We conduct experiments across different tasks and model architectures to verify the effectiveness of our approach. When combined with Adafactor as a base optimizer, our approach yields similar performance to uncompressed, full-matrix update, while largely outperforming other compression techniques such as LoRA. Interestingly, the space complexity of \name is in the same order as LoRA but has a smaller constant in practice, leading to less memory usage than LoRA. 

\section{Approach}

In this section, we first present our observation of the dynamics of LoRA updates (\S{2.1}). Then, we show that LoRA can be approximated by random projection (\S{2.2}), which serves as gradient compression (\S{2.3}) and can be used for sublinear-space gradient accumulation and momentum calculation (\S{2.4}).

\subsection{Dynamics of low-rank adaptation (LoRA)}

For update a pre-trained weight matrix $W \in \sR^{n \times m}$, LoRA parameterizes $B \in \sR^{n \times r}$ and $A \in \sR^{r \times m}$ with $r \ll \min\{n,m\}$. After applying LoRA, the forward pass becomes
\begin{align}
   y = (W + BA)x = Wx + BAx,
\end{align}
where $x \in \sR^m$ is the input for current layer and $y \in \sR^n$ is the pre-activation value of the next layer. At the beginning of LoRA updates, $BA$ should not change the original weight $W$. The common practice is to initialize the matrix $B$ with an all-zero matrix and $A$ with a normal distribution.

During back-propagation, the matrix $W$ has gradient
\begin{align}
    \nabla_W \gL =\frac{\partial L}{\partial y} x^\top,
\end{align}
where $\tfrac{\partial L}{\partial y} \in \sR^{n}$ is the partial derivative w.r.t. $y$. LoRA only calculates the gradient w.r.t. the matrices $A$ and $B$, given by
\begin{align}
    & \frac{\partial \gL}{\partial A} = B^\top \frac{\partial L}{\partial y} x^\top = B^\top (\nabla_W \gL)  \label{eq:derva} \\
    \text{and} \quad\quad & \frac{\partial \gL}{\partial B} = \frac{\partial \gL}{\partial y} x^\top A^\top = (\nabla_W \gL) A^\top. \label{eq:dervb} 
\end{align}

Our insight is that in Equations~\eqref{eq:derva} and \eqref{eq:dervb}, LoRA essentially down-projects the original gradient to a lower dimension. In fact, we found that LoRA recovers the well-known random projection method~\cite{dasgupta2000experiments, bingham2001random}. We formally state this in the following theorem.

\begin{restatable}{theorem}{lorarp}\label{thm:lora-rp}
Let LoRA update matrices $A$ and $B$ with SGD for every step $t$ by
\begin{align}
    A_{t+1} \gets& A_t - \eta B_t^\top (\nabla_W \gL_t)  \\
    B_{t+1} \gets& B_t - \eta (\nabla_W \gL_t) A_t^\top,
\end{align}
where $\eta$ is the learning rate.
We assume $ \big \|  \sum_{t=0}^{T}  \nabla_W \gL_t \big\|_F \le L$ for every $T$ during training, which implies that the model stays within a finite Euclidean ball. In this case, the dynamics of $A_t$ and $B_t$ are given by \begin{align}
      A_T = A_0 + \eta A_0 f_A(T), \quad B_T = \eta f_B(T) A_0^\top,
  \end{align}
  where the forms of $f_A(t) \in \sR^{m \times m}$ and $f_B(t) \in \sR^{n \times m}$ are expressed in the proof. In particular, $\| f_A(t) \|_2 \le \frac{\eta L^2 \big(1 - (\eta^2 L^2)^t\big)}{1-\eta^2 L^2}$ for every $t$.
\end{restatable}

\begin{proof}
    See Appendix~\ref{apx:prf-lora-approx}.
\end{proof}

Theorem~\ref{thm:lora-rp} describes the SGD dynamics of LoRA updates. Without loss of generality, we denote the total change of $A$ and $B$ after $T$ step as $\Delta A$ and $\Delta B$, respectively. Then the fine-tuned forward function will be \begin{align}
    & W + (B_0 + \Delta B)(A_0 + \Delta A) \\
    =& W + B_0 A_0 + B_0 \Delta A + \Delta B A_0 + \Delta B \Delta A  \\
    =& W + \Delta B A_0 + \Delta B \Delta A, \label{eq:exact} 
\end{align}
where $B_0=0$ is due to the initialization of the $B$ matrix. The final expression dissects the LoRA weight into two parts. We observe that it is the first part that dominates the total weight change, as stated below.

\begin{observation}\label{obs:approx}
    When the learning rate is small, we have an approximation that 
    \begin{align}
        W + (B_0 + \Delta B)(A_0 + \Delta A) \approx W + \Delta B A_0.
    \end{align}
\end{observation}

This can be seen by expanding $B_0$ and $A_0$ given by Theorem~\ref{thm:lora-rp}. Specifically, we have that \begin{align}
    & W + \Delta B A_0 + \Delta B \Delta A \\
    =& W + \eta f_B(t) A_0^\top A_0 + \eta^2 f_B(t) A_0^\top A_0 f_A(t) 
\end{align}

Our insight is that the third term has a smaller magnitude when the learning rate is not large. This is because $\|f_A(t)\|_2 \le \|f_A(t)\|_F \le \tfrac{\eta L^2 \big(1 - (\eta^2 L^2)^t\big)}{1-\eta^2 L^2}$ given by Theorem~\ref{thm:lora-rp}. If $\eta \ll 1/L$, we have $\lim_{t\to\infty} \eta \| f_A(t) \| \ll 1$, which indicates that the third term is significantly smaller than the second term, making it negligible in the final updates.

\subsection{Random projection of gradients}\label{sec:comp-decomp}

By Observation~\ref{obs:approx}, the change of the matrix $B$ dominates the final weight. A straightforward simplification is to freeze the matrix $A$ but to tune the matrix $B$ only (denoted by $\Delta \tilde B)$. In this case, we have \begin{align}
    & W + (B_0 + \Delta  B) (A_0 + \Delta A) \\
    \approx & W + \Delta \tilde B A_0 \label{eq:dropdab}\\
    =:&  W + \eta \tilde f_B(T) A_0^\top A_0. \label{eq:use-ftilde}
\end{align}
In Equation~\eqref{eq:dropdab}, $B_0$ is dropped because $B$ is initialized as an all-zero matrix. Equation~\eqref{eq:use-ftilde} defines $\tilde f_B(T)$, which will have the update form
\begin{align}
    \tilde f_B(t + 1) := \tilde f_B(t) - \nabla_W \gL_t
\end{align}
following the derivations in Theorem~\ref{thm:lora-rp}. Therefore, $\tilde f_B(t)=-\sum_i \nabla_W \mathcal L_i$. Putting it to Equation~\eqref{eq:use-ftilde}, we have
\begin{align}
    &  W + \eta \tilde f_B(T) A_0^\top A_0 \\
    =& W - \eta \left(\sum_{t=0}^T \nabla_W \gL_t \right)  A_0^\top A_0\\
    =& W - \eta \sum_{t=0}^T\Big[ (\nabla_W \gL_t) A_0^\top A_0\Big]. \label{eq:rp}
\end{align}
In other words, our derivation reveals that, with some approximations, \textbf{LoRA updates can be viewed as performing random projection to the gradient}. In particular, it compresses a gradient by a random down-projection $A_0^\top$,  and then decompresses it by an up-projection  $A_0$.

\begin{figure}[t]
    \centering
    \includegraphics[width=0.85\linewidth]{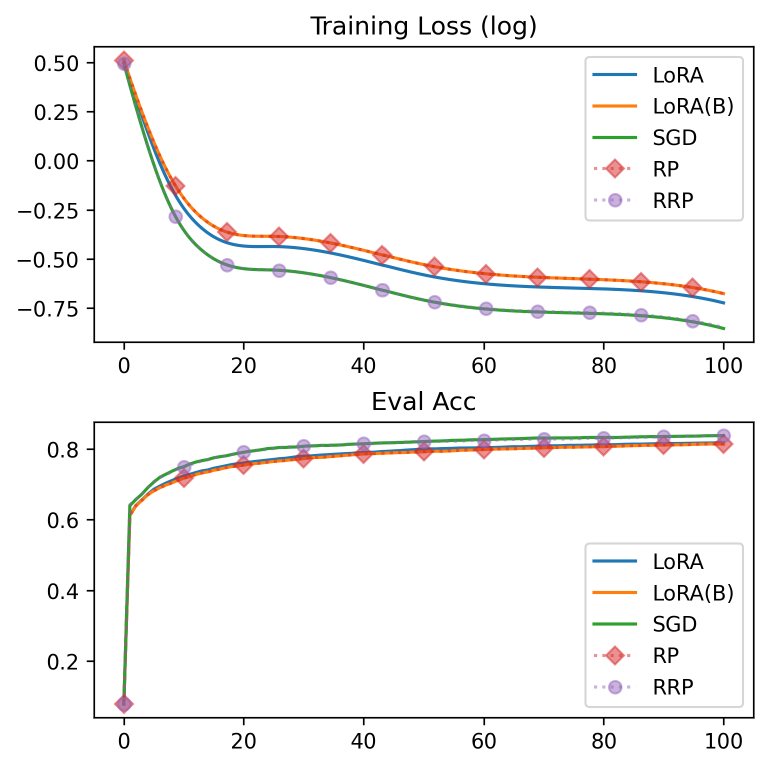}
    \caption{The results of LoRA and its simplifications. We apply the LoRA patch to the first layer of the network with a shape of $768\times768$ and set $r=8$. The legend \textit{LoRA} is the original LoRA method, while \textit{LoRA(B)} is the simplification where only the matrix $B$ is updated. \textit{RP} (random projection) and \textit{RRP} (resampled RP) follow the same update rule~\eqref{eq:rp}, but \textit{RRP} uses different projection matrices at different steps. In addition, we show the results of SGD on the full model for comparison. All experiments use the same $\eta =0.01$.}
    \label{fig:preliminary}
\end{figure}

\subsection{Our interpretation of LoRA}\label{sec:interpretation}

To this end, we provide a novel interpretation of a LoRA update by framing it as the compression and decompression of gradients.

\paragraph{Compression.} LoRA first compresses the gradient by a random down-projection, which can be justified by the following result based on the Johnson--Lindenstrauss lemma~\cite{dasgupta2003elementary, matouvsek2008variants}.

\begin{lemma}[\citeauthor{indyk1998approximate}]\label{lem:im}
    Let $\epsilon \in (0, 1/2]$ and $\delta \in (0, 1)$. Let  $A \in \sR^{r \times m}$ be a random matrix where each element is independently sampled from a standard Gaussian distribution. There exists a constant $c$ such that when $r = c \epsilon^{-2} \log (\delta/2)$, we have \begin{align}
        (1-\epsilon) \|x\| \le (1/{\sqrt{r}}) \|  Ax \| \le (1 + \epsilon) \|x\|
    \end{align}
    with probability at least $1-\delta$ for every $x \in \sR^m$.    
\end{lemma}

Essentially, this lemma suggests that, with a high probability, the projection by a random Gaussian matrix largely preserves the scaled norm in the original space.

In the case of LoRA, such a random projection is applied to each row of the gradient matrices, whose dimension is thus reduced from $\sR^{n\times m}$ to $\sR^{r\times m}$. The lemma asserts that the norm structure of the rows is approximately preserved.

\paragraph{Decompression.} After down-projection by $A_0^\top$, LoRA decomposes the gradient by an up-projection $A_0$. We show that this will recover the original gradient in expectation:
 \begin{align}
    & \E_{A_0} \left[ W + (\nabla_W \gL_t) A_0^\top A_0 \right]\\ 
    =& W + (\nabla_W \gL_t) \E_{A_0}[A_0^\top A_0]
\end{align}
where $(1/r) \E_{A_0}[A_0^\top A_0]$ is an identity matrix. Moreover, the larger the rank $r$, the closer the expectation is to the identity. We quantify the error in our following theorem.

\begin{restatable}{theorem}{decompression}\label{thm:decompression}   
Let $A$ be a matrix of shape $\sR^{r \times m}$ where each element is independently sampled from a standard Gaussian distribution. Let $\epsilon, \delta\in (0,1]$. There exists a constant $c$ such that when $r = c \log (2m/\delta) \epsilon^{-2}$, we have  for all $i,j$ that \begin{align}
    \Big| [A^\top A - I]_{i,j} \Big| \le \epsilon
\end{align} with confidence at least $1-\delta$.
\end{restatable}
\begin{proof}
    See Appendix~\ref{apx:prf-decompression}.
\end{proof}

Our Theorem~\ref{thm:decompression} implies that $r$ only needs to scale logarithmically to preserve the element-wise reconstruction error, which is efficient in both computation and memory. Further, the logarithmic asymptotic rate makes it an ideal candidate to be applied to the training of modern neural models where $m$ is large.

We empirically verify our interpretation of LoRA by a pilot study on the Fashion-MNIST dataset~\cite{xiao2017fashion} with a simple feed-forward network. We experiment with a variant of LoRA where only $B$ is tuned; we call the variant LoRA(B). As shown in Figure~\ref{fig:preliminary}, the performance of LoRA(B) is close to the original LoRA, which is consistent with Observation~\ref{obs:approx} and suggests the overall update of LoRA is dominated by the compression-and-decompression step. Further, the curve is identical to random projection (RP), well aligned with our derivation in Section~\ref{sec:comp-decomp}. 

\subsection{Our method: \name}\label{sec:approach:flora}

Based on the analyses, we propose our method, \name (\longname), to enable overall high-rank updates. One of the main insights of \name is that it constantly resamples the projection matrix in Equation~\eqref{eq:rp}. Therefore, our total weight change is no longer constrained to be low-rank. Moreover, we can apply the random-projection compression to the optimization states for memory saving. We demonstrate two common scenarios where \name can be applied: (1) an arithmetic mean (AM) over a period of history, for which a concrete example is gradient accumulation; (2) an exponential moving average (EMA), for which an example could be momentum calculation. We show that compression of \name has the same asymptotic rate as LoRA but with a lower constant.

\paragraph{Resampling random projection matrices.}

With the approximation in Observation~\ref{obs:approx}, LoRA can be viewed as having a fixed random projection matrix $A_0$. This restricts the overall change of $W$ to be low-rank. However, our analysis in Section~\ref{sec:interpretation} holds for any random matrix at every time step. 

Therefore, we propose in \name to resample a new random matrix to avoid the total change restricted in a low-rank subspace. In our pilot study, resampling the random matrix (RRP) largely recovers the performance of full-matrix SGD, significantly surpassing both the original LoRA and its approximated version in Equation~\eqref{eq:rp}. The empirical evidence highlights the effectiveness of avoiding the low-rank constraint in our \name. 

It should be emphasized that we cannot resample the down-projection matrix $A$ in LoRA. This is because $A$ and $B$ are coupled during the updates, and if the down-projection matrix $A$ is resampled, the already updated matrix $B$ will not fit. On the other hand, our \name directly updates the weight matrix $W$ during training, making it flexible to choose a random down-projection at every step.

\paragraph{Sublinear memory gradient accumulation.}

\begin{algorithm}[tb]
    \small
    \caption{Gradient accumulation with \name.}
    \label{alg:grad-accum}
 \begin{algorithmic}[1]
     \REQUIRE rank $r \in \sZ_+$, accumulating steps $\tau \in \sZ_+$
     \REQUIRE gradient function $ \nabla_W f_{\cdot}(\cdot)$
     \REQUIRE weight matrices $\sW = \left\{W^{(l)}: \dim(W^{(l)}) = 2\right\}$ 
 
      \COMMENT{Initialization of the accumulator state}
 
     \FOR{$W \in \sW$}
           \STATE {$C_{W} \gets \mathbf{0}^{n\times r}$} \hfill \COMMENT{$O(nr)$}
           \STATE {$s_{W} \gets \text{an independent random seed}$}
     \ENDFOR
     
     \COMMENT{Accumulating the compressed gradients}
     \FOR{$i \in [\tau]$}
          \FOR{$W \in \sW$}
               \STATE {$G_W \gets \nabla_W f_i(\sW)$} \hfill \COMMENT{$G_W \in \sR^{n\times m}$}
               \STATE {$A_W \gets \gN_{s_W}(0, 1/r)$} \hfill \COMMENT{$A_W \in \sR^{r\times m}$}
               \STATE {$C_W \gets C_W + G_W A_W^\top$} \hfill \COMMENT{Compresssion}
         \ENDFOR
     \ENDFOR
 
     \COMMENT{Reconstruction}
     \FOR{$W \in \sW$}
         \STATE {$A_W \gets \gN_{s_W}(0, 1/r)$} \hfill \COMMENT{$A_W \in \sR^{r\times m}$}
         \STATE {$\tilde G_W \gets (1/n) C_W A_W $}  \hfill \COMMENT{Decompression}
     \ENDFOR
     \STATE{\textbf{return}  $\{ \tilde G_W : W \in \sW \}$} \hfill \COMMENT{Overwrite $\{G_W\}$}
 \end{algorithmic}
 \end{algorithm}
 
One application of our \name is to compress the optimization states to save memory during training. We first show this with an example of gradient accumulation, which is widely used in practice to simulate a larger batch size~\cite{smith2018don}. Specifically, it calculates the arithmetic mean (AM) of gradients for $\tau$ steps and updates the model by the AM. In this way, the effective batch size is $\tau$ times larger than the original batch size. However, it requires a memory buffer, whose size is equal to the model itself, to store the accumulated gradients.

In \name, we propose to compress the gradient accumulation with the down-projection. Within an accumulation cycle, we only maintain the accumulated gradient in the randomly down-projected space. During decompression, we can reuse the memory for gradients to reduce additional overheads. The resampling of the projection matrix occurs when an accumulation cycle is finished. The overall algorithm is summarized in Algorithm~\ref{alg:grad-accum}.

\paragraph{Sublinear memory momentum.}

\begin{algorithm}[tb]
    \small
    \caption{Momentum with \name.}
       \label{alg:momentum}
    \begin{algorithmic}[1]
       \REQUIRE decay rates $0\le \beta \le 1$
       \REQUIRE rank $r \in \sZ_+$, interval $\kappa \in \sZ_+$
       \REQUIRE gradient function $ \nabla_W f_{\cdot}(\cdot)$
       \REQUIRE weight $\sW = \left\{W^{(l)}: \dim(W^{(l)}) = 2\right\}$ 
    
       \COMMENT{Initialize the optimizer state}
       \STATE{$t \gets 0$}
       \FOR{$W \in \sW$}
            \STATE {$M_{t,W} \gets \mathbf{0}^{n\times r}$}
            \STATE {$s_{t,W} \gets \text{an independent random seed}$}
       \ENDFOR
    
       \COMMENT{Training procedure}
       \WHILE{training not converged}
           \FOR{$W \in \sW$}
               \STATE {$G_{t,W} \gets \nabla_W f_t(\sW)$} \hfill \COMMENT{$G_{t,W} \in \sR^{n\times m}$}

                \STATE {$A_{t,W} \sim \gN_{s_{t,W}} (0, 1/r)$} \hfill \COMMENT{$ A_{t,W} \in \sR^{r \times m}$}

                \IF{$t \equiv 0 \pmod \kappa$}
                    \STATE {$s_{t + 1,W} \gets \text{an independent random seed}$}

                    \STATE {$A'_W \sim \gN_{s_{t+1,W}} (0, 1/r)$} \hfill \COMMENT{$ A'_W \in \sR^{r \times m}$}
                    \STATE{$M' \gets M_{t,W} A_{t,W} {A'_W}^\top $}

                \ELSE
                    \STATE {$s_{t + 1,W} \gets s_{t,W}$}
                    \STATE{$A'_W \gets A_{t,W}$}
                    \STATE {$M' \gets M_{t,W}$}
                \ENDIF

                $M_{t+1,W} \gets \beta M' + (1-\beta) G_{t,W} {A'_W}^\top $
           \ENDFOR
           \STATE{\textbf{yield} $\{M_{t+1,W} A'_W: W \in \sW\}$} \hfill\COMMENT{Decompression}
          \STATE{$t \gets t+1$}
       \ENDWHILE
    \end{algorithmic}
    \end{algorithm}

The momentum technique is widely used in modern deep learning to reduce the variance of the gradient and accelerate training~\cite{nesterov1998introductory, goh2017why, jelassi2022towards}. However, it requires maintaining an additional momentum scalar for each parameter, which is expensive when the model is large. 

Similar to the compressed gradient accumulation, we can also compress the momentum with \name. For each time step $t$, we down-project the new gradient $G_t$ by a random projection matrix $A_t^\top$. However, the difficulty for accumulating momentum emerges when we use a $A_t$ different from $A_{t-1}$, as we cannot reconstruct the original momentum from $G_{t-1} A^\top_{t-1} + G_t A_t^\top$. This difficulty applies to all EMA updates where the number of accumulation steps is not finite. In this case, resampling new matrices will result in a loss of historical accumulation.

We propose two remedies to address this issue. First, we keep the same projection matrix for a long time to reduce the distortion. Second, we propose to transfer the compressed momentum from the old projection to a new one by $M_{t}=M_{t-1} A_{t-1} A_t^\top$. This is justified by $A_{t-1}^\top A_{t-1}$ and $A_t^\top A_t$ are approximately the identity matrix based on Theorem~\ref{thm:decompression}.

The final algorithm is shown in Algorithm~\ref{alg:momentum}. Overall,  for each weight matrix $W\in\mathbb R^{n\times m}$, we preserve the momentum term $M_t$ with sublinear memory. Compared with the original momentum, we reduce the memory from $O(nm)$ to $O(nr)$. It should be pointed out that although we track momentum specifically in this case, our algorithm can be easily extended to other EMA-based statistics.

\begin{table*}[t!]
\small
\centering
\caption{The results of different methods to compress gradient accumulation. The size indicates the total number of the original model parameters. The numbers in the brackets denote the rank $r$ of the random projection matrix.}\label{tab:ga}
\begin{subtable}[t]{0.5\textwidth}
    \centering
    \caption{The results of T5 variants on XSum.}\label{tab:ga-t5}
    \begin{tabular}{lllcc}
    \toprule
    Size   & Accumulation  &  Mem & $\Delta_M$ & R$_1$/R$_2$/R$_\text{L}$  \\
    \midrule
    \multirow{10}{*}{60M}  &  None & 0.75 & - &33.4/11.4/26.4 \\ 
    & Naive & 0.87 & 0.12  &34.0/11.5/26.7 \\
    & LoRA(8) & 0.82 & 0.07 & 30.4/8.60/23.6  \\
    & LoRA(32) & 0.86 & 0.11 & 30.7/8.90/23.9 \\
    & LoRA(128) & 0.94 & 0.19 & 31.0/9.10/24.1\\
    & LoRA(256) & 1.07 & 0.32 &31.4/9.34/24.5\\
    & \textsc{Flora}(8) &  0.75 & 0.00 &31.5/9.67/24.6 \\
    & \textsc{Flora}(32) &  0.75  & 0.00 & 32.2/10.3/25.2 \\
    & \textsc{Flora}(128) & 0.77 & 0.02 &33.2/10.9/26.0\\
    & \textsc{Flora}(256) & 0.79 & 0.04 &33.6/11.3/26.5  \\
    \midrule
    \multirow{10}{*}{3B}  &  None & 16.7 & - &42.5/19.1/34.6   \\ 
    & Naive & 26.6 & 9.9 &44.4/20.9/36.3 \\
    & LoRA(16) & 27.8 & 11.1 &42.2/18.4/34.0 \\
    & LoRA(64) & 29.5 & 12.8 &42.3/18.6/34.1   \\
    & LoRA(256) & 33.4 & 16.7 &42.6/18.9/34.4  \\
    & LoRA(512) & OOM & -&- \\
    & \name(16) &  17.0 & 0.3 &  43.5/20.0/35.5 \\
    & \name(64) &  18.2  & 1.5 & 43.9/20.3/35.8 \\
    & \name(256) & 19.5  & 2.8 &44.3/20.7/36.2 \\
    & \name(512) &  22.1  & 5.4 &44.5/20.9/36.4 \\
    \bottomrule
    \end{tabular}
\end{subtable}
\begin{subtable}[t]{0.45\textwidth}
    \centering
    \caption{The results of GPT-2 variants on IWSLT17 De-En.}\label{tab:ga-gpt}
    \begin{tabular}{lllcc}
    \toprule
    Size   & Accumulation  &  Mem & $\Delta_M$ & BLEU  \\
    \midrule
    \multirow{10}{*}{110M}  & None & 2.77 & - & 17.9 \\ 
        ~ & Naive & 3.24 & 0.47 & 24.9 \\ 
        ~ & LoRA(8) & 3.25 & 0.48 & 9.94 \\ 
        ~ & LoRA(32) & 3.29 & 0.52 & 11.2\\ 
        ~ & LoRA(128) & 3.38 & 0.60 & 12.2  \\ 
        ~ & LoRA(256) & 3.52 & 0.75 & 13.4 \\ 
        ~ & \name(8) & 2.93 & 0.15 & 16.3 \\ 
        ~ & \name(32) & 2.94 & 0.16 & 22.0 \\ 
        ~ & \name(128) & 2.98 & 0.20 & 24.0 \\ 
        ~ & \name(256) & 3.03 & 0.26 & 25.4 \\
    \midrule
    \multirow{10}{*}{1.5B}
        & None & 20.8 & - & 28.2 \\ 
        & Naive & 26.5& 5.78 & 33.2 \\ 
        & LoRA(16) & 26.8 & 6.02 & 17.4 \\ 
        & LoRA(64) & 27.4 & 6.68 & 19.5 \\ 
        & LoRA(256) & 28.9 & 8.15 & 20.7 \\ 
        & LoRA(512) & OOM & - & - \\ 
        & \name(16) & 21.1 & 0.34 & 29.7 \\ 
        & \name(64) & 21.3 & 0.52 & 31.6 \\ 
        & \name(256) & 21.9 & 1.17 & 33.2 \\ 
        & \name(512) & 22.8 & 2.04 & 33.6 \\ 
    \bottomrule
    \end{tabular}
\end{subtable}
\end{table*}

\paragraph{Memory analysis.}

It should be pointed out that neither LoRA nor our \name saves the memory for back-propagation. This is because $\tfrac{\partial \gL}{\partial W}$ is needed for the update of $A$ and $B$ in LoRA~\cite{dettmers2023qlora}, while in our approach we also compress and decompress the gradient.

That being said, saving the memory of optimization states alone could be critical to training large models~\cite{dettmers20218bit}. Our \name compresses both the AM and EMA of gradients to the sublinear level, which shares the same asymptotic rate as LoRA. In implementation, we may store the random seed that generates the projection matrix---which is highly efficient, as each element can be sampled independently with simple operations---instead of maintaining the same project matrix over batches. This allows the program to further save memory in practice with buffer reuse. On the contrary, LoRA needs to maintain two weight matrices $A$ and $B$, as well as their AM or EMA matrices. Empirically shown in Section~\ref{sec:exp}, \name consumes less memory than LoRA while facilitating high-rank updates and outperforming LoRA to a large extent.

\section{Experiments}\label{sec:exp}

In this section, we empirically verify the effectiveness of our approach across different model architectures and datasets.

\subsection{Experiment setup}

\paragraph{Models.}
Given the exceptional ability of language models, we consider Transformer-based models for our experiments. Specifically, we select two representative models, including the T5~\cite{raffel2020exploring} and GPT-2~\cite{radford2019language} series. For the T5 series, we use T5-small to represent small models and T5-3B to represent large models. T5-small has around 60M parameters, with a hidden dimension set to 512, while T5-3B has around 3B parameters with a hidden dimension of 1024. For the GPT-2 series, we use the base version to represent small models and GPT-2-XL to represent large models. The base version has around 110M parameters, with a hidden dimension set to 768, while the large version has around 1.5B parameters with a hidden dimension of 1600.

\paragraph{Datasets.}

\begin{table}[t]
    \small
    \centering
    \caption{The results of compressing momentum. The size indicates the total number of the original model parameters. The numbers in the brackets denote the rank $r$ of the random projection matrix.}\label{tab:mom}
    \begin{tabular}{llcc}
    \toprule
    Setting   & Momentum  &  Mem & Results  \\
    \midrule
    \multirow{10}{*}{\shortstack[l]{T5 60M \\ XSum}}  & None & 1.65 & 29.4/9.11/23.3 \\ 
    & Naive & 1.89 &  29.9/9.40/23.8 \\
    & LoRA(8) & 1.88  & 18.0/3.33/14.9 \\
    & LoRA(32) & 1.91  & 20.4/4.20/16.7/ \\
    & LoRA(128) & 2.05 & 21.5/4.82/17.4 \\
    & LoRA(256) & 2.13 & 22.2/5.04/17.9 \\
    & \name(8) &  1.71 &  25.5/6.56/20.4 \\
    & \name(32) &  1.72  & 26.9/7.32/21.5  \\
    & \name(128) & 1.75  & 29.1/8.76/23.2 \\
    & \name(256) & 1.79  &  30.2/9.51/24.0  \\
    \midrule
    \multirow{10}{*}{\shortstack[l]{GPT-2 110M \\ IWSLT17}}  &  None & 8.95 &  19.4 \\ 
    & Naive & 9.45 & 19.9 \\
    & LoRA(8) & 9.42 & 4.98 \\
    & LoRA(32) &  9.46& 6.76 \\
    & LoRA(128) &  9.55 & 8.72 \\
    & LoRA(256) & 9.76 & 9.83 \\ 
    & \name(8) & 9.09 &  9.14 \\
    & \name(32) &  9.10 &  14.9 \\
    & \name(128) &  9.14 & 18.6 \\
    & \name(256) & 9.20 & 19.9 \\
    \bottomrule
    \end{tabular}
\end{table}

To facilitate evaluation, we use two conditional language modeling tasks, including summarization and translation.

For the summarization task, we train\footnote{In our experiments, we fine-tune a pre-trained model in the gradient accumulation experiment, while training from scratch in the momentum experiment (Section~\ref{sec:exp-results}).} T5 on the XSum dataset~\cite{narayan2018don}. Each sample is a news article with a summary. The task is to generate a summary of the article. For each input, we prepend the prefix ``summarize:'' to the source sentence in the encoder~\cite{raffel2020exploring}. The source and target sentences are truncated to 512 and 128 tokens, respectively.

For the translation task, we follow the setting of \citet{lin2020exploring} and train GPT-2 on the IWSLT-2017 German-English dataset~\cite{cettolo2017overview}. Each sample is a German sentence with its English translation. The task is to generate the English translation of the German sentence. For each input, we use the template ``translate German to English: \texttt{[source]}. English: \texttt{[target]}'' for training~\cite{raffel2020exploring}. 

\paragraph{Evaluation metrics.}
For the summarization task, we use the widely used ROUGE scores~\cite{lin2004rouge}, including ROUGE-1, ROUGE-2, and ROUGE-L (R$_1$/R$_2$/R$_\text{L}$) to evaluate the quality of the generated summary. For the translation task, we use the most commonly used SacreBLEU score~\cite{post2018call} to evaluate the translation quality. For both ROUGE and SacreBLEU scores, the higher the score, the better the quality of the generated text. 

To get more insights into the training process, we monitor the peak memory usage with the built-in JAX profiling tool~\cite{jax2018github}. We also show the excessive memory $\Delta_M$ compared with the method where accumulation or momentum is disabled. The memory is reported in GiB ($1024^3$ bytes).

\paragraph{Competing methods.} In our experiment, we take Adafactor as the base optimizer, which is the default optimizer for many Transformer models including T5~\cite{raffel2020exploring}, and is reported to be empirically better than Adam~\cite{rae2021scaling}. We use the official Adafactor implementation in Optax~\cite{deepmind2020jax}.

We compare the following methods: (1) \textbf{None}: a baseline that does not use gradient accumulation or momentum; (2) \textbf{Naive}: a naive implementation of gradient accumulation or momentum, which stores the full information along training; (3) \textbf{LoRA}: the original LoRA method where only the LoRA patches are trained; (4) \textbf{\name}: our approach that compresses the gradients and decompresses them when updating the original weights. For LoRA and \name, we apply the projections to attention and feed-forward layers only, while following the naive procedure for other layers (i.e., token embeddings and vector weights). 

For small models (T5-small and GPT-2 base), we test the rank $r$ from 8 to 256, ranging from the very low dimension to half of the hidden dimension, for a thorough examination of different methods. For large models (T5-3B and GPT-2-XL), we test $r$ from 16 to 512 to approximately maintain the same percentage of memory saving as small models. We do not apply learning rate schedules~\cite{loshchilov2017sgdr} or weight decay~\cite{loshchilov2018decoupled} in any experiments to rule out the influence of these techniques. 

\subsection{Main results}\label{sec:exp-results}
\paragraph{Gradient accumulation.}

In this setting, we fine-tune pre-trained models with 16 gradient accumulation steps. To achieve a minimal memory footprint and fit large models,  the physical batch size is set to 1.  We sweep the learning rate from~$10^{-5}$ to~$10^{-1}$ with the naive accumulation method on the validation loss. The best learning rate is applied to other methods, excluding LoRA, which is tuned individually as it is reported to have different optimal learning rates~\cite{hu2022lora}. For each run, the model is fine-tuned for 1 epoch to prevent over-fitting following the common practice~\cite{wu2021recursively}. The results are reported on the test set based on the checkpoint with the lowest validation loss.

We present the results in Table~\ref{tab:ga}. As shown, the naive gradient accumulation improves the ROUGE scores over the method without accumulation, but it leads to a large memory usage, which is similar to the model size, to store the accumulation.  For LoRA, we empirically observe that it generally does not reduce memory usage in this case, as the state of Adafactor is already sublinear. In fact, it increases memory because it stores another four low-rank matrices for each weight matrix and adds an additional Jacobian path for the automatic differentiation.

On the contrary, our  \name reduces the memory footprint on all benchmarks compared with the naive accumulation. In addition, when $r$ is reasonably large, our method is able to recover the performance (in ROUGE or BLEU scores) of full-matrix accumulation and surpass the baseline that accumulation is not enabled. Notably, for the large models (T5-3B and GPT-2-XL), the memory overhead of \name ($r=256$) is only $30\%$ of the naive accumulation, while the performance is on par.

\paragraph{Momentum.}
Given that the momentum technique is ineffective in fine-tuning~\cite{li2020rethinking}, we train all models from scratch in this setting. The physical batch size is set to $4$ as a result of balancing the generalization and variance reduction~\cite{masters2018revisiting}. We disable the gradient accumulation technique to rule out its impact. Due to the expense of training from scratch, we only test the small variants of each series.  Similar to the settings in gradient accumulation, we sweep the learning rate for the naive momentum method from~$10^{-5}$ to~$10^{-1}$ on the validation loss.  The best learning rates are applied to all methods excluding LoRA, which again has its own optimal learning rate. The hyper-parameter $\kappa$ (resampling interval) is set to $1000$ for all runs of \name. The effect of different values of $\kappa$ is shown in Section~\ref{sec:analyses}.

As shown in Table~\ref{tab:mom}, the naive momentum technique achieves better performance than at a cost of more memory usage. Similar to the results in gradient accumulation, LoRA does not save memory given the optimization state is already sublinear. It also has a significantly lower performance when trained from scratch, as the overall matrix update can only be low-rank.

Our \name utilizes less memory than the naive momentum. In addition, our method recovers (or even surpasses) the performance of naive momentum when $r$ is increased. This significantly distinguishes our methods from LoRA as it achieves memory-efficient training even when the initialization is random.

\subsection{In-depth analyses}\label{sec:analyses}

\paragraph{The effect of $\kappa$ in momentum.}

In our momentum implementation, we have a hyper-parameter $\kappa$ controlling the resampling frequency of the random-projection matrix. In this part, we analyze the effect of $\kappa$ with T5-small on the summarization task as our testbed, due to the limit of time and resources. We vary $\kappa$ by keeping other hyper-parameters the same as in Section~\ref{sec:exp-results}.

\begin{table}[t]
    \small
    \centering
    \caption{The effect of $\kappa$ in momentum.}\label{tab:kappa}
    \begin{tabular}{llcc}
    \toprule
    Setting   & $\kappa$  &  Mem &  R$_1$/R$_2$/R$_\text{L}$   \\
    \midrule
    \multirow{5}{*}{\shortstack[l]{T5 60M \\ XSum}}   & 1 & 1.79 & 0.00/0.00/0.00 \\
    & 10& 1.79 & 27.5/7.68/31.8\\
    & 100 & 1.79 & 29.3/8.89/23.2 \\
    & 1000 & 1.79 & 30.4/9.70/24.2 \\
    & 10000 & 1.79 & 29.5/9.11/23.5 \\
    \bottomrule
    \end{tabular}
\end{table}

The results are shown in Table~\ref{tab:kappa}. It is seen that, when $\kappa$ is below $1000$, the ROUGE scores increase with $\kappa$. After a certain threshold, however, we see that the performance starts to decrease. This aligns with our interpretation that the information is better preserved within the interval, but each interval is bottlenecked by the rank. Given the results, we choose $\kappa=1000$ in Section~\ref{sec:exp-results} to balance the preserved information and the overall rank of momentum.

\paragraph{Optimizer with linear memory.}

In our main experiments, we observed a counter-intuitive phenomenon that LoRA empirically increases memory usage. This is likely because the optimization states in Adafactor are already sublinear, rendering the ineffectiveness of LoRA to save memory in this case. To further verify our method in linear-memory optimizers, we test the performance with a variant of Adafactor where the second-moment estimates are not factorized, essentially making it a linear-memory optimizer. All the other hyper-parameters remain the same as Section~\ref{sec:exp-results}.

\begin{table}[t]
    \small
    \centering
    \caption{The results of linear-memory optimizers.}\label{tab:linear-opt}
    \begin{tabular}{llcc}
    \toprule
    Setting   & Accumulation  &  Mem &  R$_1$/R$_2$/R$_\text{L}$   \\
    \midrule
    \multirow{10}{*}{\shortstack[l]{T5 60M \\ XSum}}  &  None & 0.99  & 33.0/11.1/26.1 \\ 
    & Naive & 1.12 & 34.0/11.5/26.7 \\
    & LoRA(8)& 0.82 & 28.7/7.51/22.0\\
    & LoRA(32) & 0.86 &  29.0/7.71/22.3\\
    & LoRA(128) & 1.00 & 29.7/8.02/22.9 \\
    & LoRA(256) & 1.20 & 30.0/8.28/23.2 \\
    & \name(8) & 1.00  & 31.6/9.72/24.7 \\
    & \name(32) & 1.00  &  32.3/10.3/25.3\\
    & \name(128) &  1.00 & 33.2/10.9/26.0 \\
    & \name(256) & 1.04 & 33.5/11.1/26.3 \\
    \bottomrule
    \end{tabular}
\end{table}

Table~\ref{tab:linear-opt} shows the results. As seen, LoRA indeed saves more memory than our \name when the rank is low ($r<$128) in linear-memory optimizers. However, our \name becomes more memory-efficient for $r=256$, because we have a lower constant in the complexity. Moreover, our  \name largely outperforms LoRA in all settings by 2--3 ROUGE points, showing the superiority of our approach.

\subsection{Additional experiments}

We additionally conducted several preliminary experiments during the author response phase and found the following observations: (1)  \name performs well on images with ViT (Appendix~\ref{apx:exp:modal}), (2) \name outperforms the concurrent GaLore in both memory reduction and model quality (Appendix~\ref{apx:exp:galore}), and (3)  \name can be combined with activation checkpointing (AC) and layer-wise update (LOMO) to further reduce the peak memory (Appendix~\ref{apx:exp:prof}).

\section{Related work and discussion}\label{sec:related}

\paragraph{Parameter-efficient fine-tuning.}

Many methods have been proposed to improve the parameter efficiency of fine-tuning large models. A straightforward way is to tune a subset of the model, such as the top layers~\cite{li2021prefix} and bias vectors~\cite{zaken2022bitfit}. Another way is to add small tunable modules, e.g., the Adapter~\cite{houlsby2019parameterefficient} and LoRA~\cite{hu2022lora}, to the pre-trained model. Although reducing the optimization memory, these methods suffer from the problem that the model parameters are restricted. For example, the total weight change of LoRA is constrained to be low-rank. In an attempt to achieve high-rank updates, ReLoRA~\cite{lialin2023stack} proposes to periodically reinitialize the LoRA patch. However, it requires full-weight pre-training to work properly, growing the peak memory linearly in model size. We hence do not include ReLoRA as a sublinear-memory baseline. By contrast, our method is able to start directly from scratch and achieve the full training performance while maintaining a sublinear complexity throughout the process.

\paragraph{Matrix compression.}

Our method is closely connected to matrix compression techniques. For example, principal component analysis~\cite{shlens2014tutorial} or matrix sketching~\cite{liberty2013simple,rothchild2020fetchsgd} use singular value decomposition (SVD) to approximate the large matrix with smaller matrices. However, the SVD procedure is computationally expensive and difficult to parallelize, making it impractical for large-scale training. Another way to compress the matrix is to use random projection~\cite{bingham2001random}, which lies the foundation of our method. Our method additionally involves a simple and efficient decompression procedure justified by Theorem~\ref{thm:decompression}. The simplification saves both computation and memory usage.

\paragraph{Memory-efficient optimizers.}

Optimization states contribute significantly to memory usage for large-scale training~\cite{dettmers20218bit}. Memory-efficient optimizers~\cite{shazeer2018adafactor,feinberg2023sketchy} are shown to effectively reduce the memory footprint. Our method is orthogonal to these methods, as it can be applied to enhance existing optimizers by compressing the momentum or gradient accumulation.

\paragraph{Memory-efficient automatic differentiation.} It is also possible to reduce the memory footprint of back-propagation with advanced techniques like activation checkpointing~\cite{chen2016training}, layer-by-layer  updating~\cite{lv2023full}, mixed-precision training~\cite{micikevicius2018mixed},  randomized auto differentiation~\cite{oktay2020randomized}, or zeroth-order optimization~\cite{malladi2023finetuning}. Technically, \name can be combined with these methods to further save memory. We demonstrate the combination of \name, activation checkpointing, and layer-by-layer updating in Appendix~\ref{apx:exp:galore} as an example. We leave further exploration to future work, given our focus in this paper is to compress optimization states.

\section{Conclusion}

\paragraph{Summary.} In this work, we introduce \name, a method based on random projections that achieves sublinear memory usage for gradient accumulation and momentum. In addition, our approach effectively addresses the low-rank limitation of LoRA by resampling projection matrices. Experimental results demonstrate significant memory reduction with maintained model performance, highlighting the potential of random projection in deep learning.

\paragraph{Future work.} 

In this paper, the largest model is 7B. For extremely large models like GPT-3, we estimate that the compressed optimization state of $r=256$ is only 2.08\% of its original memory, which would be of great practical significance. We leave the empirical verification to future work. Further, the applicable scope of \name is not limited to Transformer-based models. We would like to test it on more architectures.

\section*{Impact statement}

Our paper presents a memory-efficient method for deep learning to advance the field of machine learning. It is not expected to have direct consequences on the general public. We, however, anticipate it to have a positive impact on the environment by reducing the resources of model training.

\section*{Acknowledgments}

We thank all reviewers for their insightful comments. The research is supported in part by the Natural Sciences and Engineering Research Council of Canada (NSERC), a Mitacs Accelerate project, the Amii Fellow Program, the Canada CIFAR AI Chair Program, an Alberta Innovates Program, and the Digital Research Alliance of Canada (\href{https://alliancecan.ca}{alliancecan.ca}).

\bibliography{main}
\bibliographystyle{icml2024}

\newpage
\appendix
\onecolumn

\section{Proof of Theorem~\ref{thm:lora-rp}}\label{apx:prf-lora-approx}

\lorarp*

Before proving the theorem, we need to obtain the form of $f_A(t)$ and $f_B(t)$. We derive them in the following lemma.

\begin{lemma}\label{lem:form-f}
    When $t=0$, $f_A(0) = f_B(0) = 0$. For $t \ge 1$, the values of $f_A(t)$ and $f_B(t)$ are iteratively obtained by:
    \begin{align}
        f_A(t) &= - \eta \sum_{i=0}^{t-1} f_B^\top(i) (\nabla_W \gL_i) \\
        f_B(t) &= - \sum_{i=0}^{t-1} (\nabla_W \gL_i) (\eta f_A^\top(i) + I) 
    \end{align}
\end{lemma}

\begin{proof}
    We prove this by induction. For the base case $t=0$, it is straightforward to show $f_A(0) = f_B(0) = 0$.

    Assume $A_t = A_0 + \eta A_0 f_A(t)$ and $B_t = \eta f_B(t) A_0^\top$ holds with such functions $f_A$ and $f_B$ for $1 \dots t$. Then for $t + 1$, we have \begin{align}
        A_{t+1} =& A_t - \eta B_t ^\top (\nabla_W \gL_t) \\
        =& A_0 +  \eta A_0 f_A(t) - \eta^2 A_0 f_B^\top(t) (\nabla_W \gL_t) \\
        =& A_0 + \eta A_0 \left(f_A(t) - \eta f_B^\top(t) (\nabla_W \gL_t)\right) \\
        =& A_0 + \eta A_0 \left(f_A(t + 1)  \right),
    \end{align}
    where we have $f_A(t+1) = f_A(t) - \eta f_B^\top(t) (\nabla_W \gL_t) = - \eta \sum_{i=0}^{t} f_B^\top(i) (\nabla_W \gL_i)$ in the last line.

    Similarly, we have \begin{align}
        B_{t+1} =& B_t - \eta (\nabla_W \gL_t) A_t^\top \\
        =& \eta f_B(t) A_0^\top - \eta (\nabla_W \gL_t) (\eta f_A^\top(t) + I) A_0^\top \\
        =& \eta \left(f_B(t) -  (\nabla_W \gL_t) (\eta f_A^\top(t) + I)  \right) A_0^\top \\
        =& \eta f_B(t + 1) A_0^\top,
    \end{align}
    where we have $f_B(t+1) = f_B(t) - (\nabla_W \gL_t) (\eta f_A^\top(t) + I) = -\sum_{i=0}^{t} (\nabla_W \gL_i) (\eta f_A^\top(i) + I)$ in the last line.
\end{proof}

\begin{proof}[Proof of Theorem~\ref{thm:lora-rp}]
    Define $a_t := \frac{\eta L^2 \big(1 - (\eta^2 L^2)^t\big)}{1-\eta^2 L^2}.$
    We prove $\| f_A(t) \| \le a_t$ by induction. For the base case $t=0$, it is trivial to see $\|f_A(0)\| = 0 \le a_0$. We then assume $\| f_A(i) \| \le a_i$ for $i \le t-1$. Since $a_t$ is monotonically increasing, we know $\| f_A(i) \| \le a_{t-1}$ for $i \le t-1$.
    By using Lemma~\ref{lem:form-f}, we have \begin{align}
        f_A(t) 
        =& - \eta \sum_{i=0}^{t-1} f_B^\top(i) (\nabla_W \gL_i) \\
        =& - \eta \sum_{i=1}^{t-1} \sum_{j=0}^{i-1}  (\eta f_A(j) + I) (\nabla_W \gL_j)^\top   (\nabla_W \gL_i).
    \end{align}

    Taking the norm, we have \begin{align}
        \|f_A(t)\|_F =&  \left\| \eta \sum_{i=1}^{t-1} \sum_{j=0}^{i-1}  (\eta f_A(j) + I) (\nabla_W \gL_j)^\top   (\nabla_W \gL_i) \right\|_F \\
        \le& \eta^2  \left\| \sum_{j=0}^{t-2}  (f_A(j)) (\nabla_W \gL_j)^\top  \sum_{i=j+1}^{t-1}  (\nabla_W \gL_i) \right\|_F +  \eta \left\| \sum_{i=1}^{t-1} \sum_{j=0}^{i-1} (\nabla_W \gL_j)^\top   (\nabla_W \gL_i) \right\|_F \\
        \le& \eta^2 L  \left\| \sum_{j=0}^{t-2}  (f_A(j)) (\nabla_W \gL_j)^\top \right\|_F + \eta L^2 \tag{Lemma~\ref{lem:found-second-term}}\\
        \le& \eta^2 L^2 a_{t-1} + \eta L^2 \tag{Lemma~\ref{lem:bound-first-term}}\\
        =& \eta^2  L^2 \frac{\eta L^2  - (\eta^2 L^2)^{t} }{1-\eta^2 L^2} +  \eta L^2 \\
        =& \frac{\eta L^2   - (\eta^2 L^2)^{t+1} }{1-\eta^2 L^2}  \\
        =& a_t.
    \end{align}

    Therefore, we have $\| f_A(t) \| \le \| f_A(t) \|_F \le a_t$ for every $t$.
\end{proof}

\begin{lemma}~\label{lem:found-second-term}
    If $\| \sum_{k=0}^{t-1} (\nabla_W \gL_j)^\top \|_F \le L$, we have \begin{align}
            \left\| \sum_{i=1}^{t-1} \sum_{j=0}^{i-1} (\nabla_W \gL_j)^\top   (\nabla_W \gL_i) \right\|_F \le L^2
    \end{align}
    for every $t$.
\end{lemma}
\begin{proof}

Let $G(k) := \nabla_W \gL_k$ for simplicity. Taking the square,
\begin{align}
     & \left\| \sum_n \sum_{m=1}^n G(m)^\top G(n) \right\|_F^2 \\
    =& \sum_{i,j} \sum_n \sum_{m=1}^n [G(m)^\top G(n) ]_{i,j}^2 \\
    =& \sum_{i,j} \sum_n \sum_{m=1}^n   \Big(\sum_k  [G(m)]_{k,i} [G(n)]_{k,j} \Big)^2\\
    \le & \Big(\sum_m  \sum_{i} \sum_k  [G(m)]_{k,i}^2  \Big) \Big(\sum_{m=n}^{t-1} \sum_j \sum_k  [G(n)]_{k,j}^2\Big) \tag{Cauchy-Schwarz} \\
    \le&  \bigg(\sum_m  \sum_{i} \sum_k  [G(m)]_{k,i}^2 \bigg)^2 \\
    \le & \bigg(\sum_{i}   \sum_k \Big(\sum_m [G(m)]_{k,i} \Big)^2 \bigg) ^2 \\
    =& \Big\| \sum_m G(m) \Big\|_F^4 \\
    \le & L^4,
\end{align}
 which completes the proof by taking square roots on both sides.
\end{proof}

\begin{lemma}\label{lem:bound-first-term}
        If $\| f_A(k) \|_F \le a_k$ for all $k < t$, we have \begin{align}
            \left\| \sum_{k=0}^{t-1}  f_A(k) (\nabla_W \gL_k)^\top \right\|_F \le a_{t-1} L
        \end{align}
        for every $t$.
\end{lemma}
\begin{proof}
    Let $A(k) := f_A(k)$ and $G(k) = (\nabla_W \gL_k)$ for simplicity. Taking the square,
    \begin{align}
          & \Big\| \sum_{k=0}^{t-1}  A(k) G(k)^\top \Big\|_F^2 \\
       =& \sum_{k=0}^{t-1} \sum_i \sum_j [A(k)G(k)^\top]^2_{i,j}  \\
      =& \sum_{k=0}^{t-1} \sum_i \sum_j \Big(\sum_l [A(k)]_{i,l} [G(k)]_{j,l}\Big)^2  \\
         =& \sum_{k=0}^{t-1}\Big(\sum_i \sum_l [A(k)]_{i,l}^2\Big) \Big(  \sum_j  \sum_l  [G(k)]_{j,l}^2 \Big) \tag{ Cauchy–Schwarz} \\
  \le& \Big( \max_{0\le k < t}  \sum_i \sum_l [A(k)]_{i,l}^2\Big) \sum_{k=0}^{t-1} \Big(\sum_j  \sum_l [G(k)]_{j,l}^2\Big) \\
  \le&  \Big( \max_{0\le k < t}  \sum_i \sum_l [A(k)_{i,l}]^2\Big) \Big( \sum_j  \sum_l (\sum_{k=0}^{t-1} [G(k)]_{j,l})^2\Big) \\
    =&  \Big( \max_{0\le k < t} \| A(k) \|_F^2 \Big) \Big\|\sum_{k=0}^{t-1} G(k) \Big\|_F^2 \\
 \le & a_{t-1}^2 L^2,
 \end{align}
 which completes the proof by taking square roots on both sides.
 \end{proof} 

\section{Proof of Theorem~\ref{thm:decompression}}\label{apx:prf-decompression}

\decompression*

\begin{proof}
    For each element of $A^\top A$, we have \begin{align}
        [ A^\top A]_{i,j} = \begin{cases}
           \sum_{k=1}^r  a_{k,i}^2, &\text{ if } i = j, \\
           \sum_{k=1}^r a_{k,i} a_{k,j}, &\text{ otherwise, }
        \end{cases}
    \end{align}
    where each element $a_{i,k}$ is an independent random variable following $\gN(0, 1)$.

    For $z_{i,i} := \sum_{k=1}^r  a_{k,i}^2$, it follows the $\chi^2(r)$ distribution. By the standard Laurent-Massart bounds~\cite{laurent2000adaptive}, we obtain \begin{align}
        \delta_{i,i} := \Big| \frac{z_{i,i}}{r} - 1 \Big| \le 2\sqrt{\frac{\log(2/\delta_1')}{r}}+2 \frac{\log(2/\delta_1')}{r} \label{eq:diag-delta}
    \end{align}
    with probability at least $1-\delta_1'$.

    For $z_{i,j} := \sum_{k=1}^r a_{k,i} a_{k,j}$ where $i \not= j$, we can rewrite it as $z_{i,j} = \sum_{k=1}^r [(\frac{a_{k,i} + a_{k,j}}{2})^2  - (\frac{a_{k,i} - a_{k,j}}{2})^2]$. In addiiton, all $(\frac{a_{k,i} + a_{k,j}}{2})^2$ and $(\frac{a_{k,i} - a_{k,j}}{2})^2$ are i.i.d. $\chi^2(1)$ distributions. Define \begin{align}
        z^+_{i,j}  := \sum_{k=1}^r \bigg(\frac{a_{k,i} + a_{k,j}}{2}\bigg)^2  \quad \text{and} \quad
        z^-_{i,j} := \sum_{k=1}^r \bigg(\frac{a_{k,i} - a_{k,j}}{2}\bigg)^2,
    \end{align}
    it is easy to see that $z^+_{i,j}$ and $z^-_{i,j}$ are i.i.d. $\chi^2(r)$ distributions. In addition, $z_{i,j} = z^+_{i,j} - z^-_{i,j}$. Therefore, we have

    \begin{align}
        \delta_{i,j} :=& \Big| \frac{z_{i,j}}{r} \Big| = \bigg| \Big( \frac{z_{i,j}^+}{r}-1\Big) - \Big(\frac{z_{i,j}^-}{r} -1 \Big) \bigg| \\
        \le& \Big| \frac{z_{i,j}^+}{r} - 1\Big| + \Big| \frac{z_{i,j}^-}{r} - 1\Big| \\
        \le& 4\sqrt{\frac{\log(4/\delta_2')}{r}}+4 \frac{\log(4/\delta_2')}{r} \label{eq:off-diag-delta}
    \end{align}
    with probability at least $1-\delta_2'$.
    
    By using a union bound upon Equations~\eqref{eq:diag-delta} and \eqref{eq:off-diag-delta}, we can obtain \begin{align}
        \delta_{i,j} \le& 4\sqrt{\frac{2\log(2m/\delta)}{r}}+ 4 \frac{2\log(2m/\delta)}{r} 
    \end{align}
    for all $i,j$ with probability at least $1-\delta$. Under this condition, we further have \begin{align}
        \delta_{i,j} =& 4\sqrt{\frac{2\log(2m/\delta)}{ 128  \log (2m/\delta) \epsilon^{-2}}}+ 4 \frac{2 \log(2m/\delta)}{128 \log (2m/\delta) \epsilon^{-2}} \tag{Let $r = 128  \log (2m/\delta) \epsilon^{-2}$}\\
        =& \frac{1}{2} (\epsilon + \epsilon^2) \\
        \le& \epsilon, \tag{$\epsilon^2 \le \epsilon \le 1$}
    \end{align}
    concluding the proof by noticing $\delta_{i,j} = \Big| [A^\top A - I]_{i,j} \Big|$.
\end{proof}

\section{Additional results}~\label{apx:exp}

\subsection{Beyond the text modality}~\label{apx:exp:modal}

Although our main experiments were conducted on datasets with text generation, our method \name can actually be applied to the matrix multiplication in different architectures. To verify the performance on other modalities beyond text, we evaluate the performance with the Vision Transformer (ViT)~\cite{dosovitskiy2020image} on the CIFAR-100~\cite{krizhevsky2009learning} image classification dataset.

\begin{table}[h!]
    \centering
    \caption{The image classification results on CIFAR-100 with ViT model.}\label{tab:vit}
    \begin{tabular}{llcc}
    \toprule
        Model size & Optimizer & Accuracy & Memory \\
        \midrule
        \multirow{2}{*}{\shortstack[l]{Base}}& Adam & 91.93 & 4.12 GiB \\
         & \name & \textbf{92.15} & \textbf{3.14 GiB} (-23.8\%) \\ 
        \midrule
        \multirow{2}{*}{\shortstack[l]{Large}} & Adam & 92.97 & 8.57 GiB \\ 
         & \name & \textbf{92.98} & \textbf{5.79 GiB} (-32.4\%) \\ 
        \bottomrule
    \end{tabular}
\end{table}

The results are shown in Table~\ref{tab:vit}. Consistent with main experiments, our \name saves as much as $32.4\%$ of training memory without sacrificing the accuracy. This confirms that \name is agnostic to model architectures and domains, demonstrating the generality of our approach.

\subsection{Comparison with GaLore}\label{apx:exp:galore}

Contemporary to our work, another optimizer,  GaLore~\cite{zhao2024galore}, adapts similar down- and up-projections for memory efficiency. The key difference between \name and GaLore is that \name randomly generates the projection matrices on the fly, whereas GaLore performs SVD operations and stores the matrices on the device. To understand their empirical performance, we evaluate both methods on the language modelling pre-training task using the C4 dataset~\cite{raffel2020exploring}. We apply the same hyper-parameters as suggested in the original GaLore paper~\cite{zhao2024galore} for both methods, except that the learning rate is 3 times smaller for \name. The model architectures are adapted from Llama-2~\cite{touvron2023llama} but modified to contain target parameter sizes.

\begin{table}[t]
\centering
\caption{The results of language modeling on the C4 dataset with \name and GaLore. ``PPL'' is the token-level perplexity. The lower the PPL, the better. For the large model (7B), we only report the memory usage with a smaller batch size of 16 as the training takes months to complete.}\label{tab:galore}
\begin{tabular}{llcc}
    \toprule
        Model size & Optimizer & PPL & Memory \\
        \midrule
        \multirow{2}{*}{\shortstack[l]{60 M}} 
        & GaLore & 34.64 & 27.7 \\
         & \name & \textbf{32.52} & \textbf{27.5} \\
        \multirow{2}{*}{\shortstack[l]{350 M}} 
        & GaLore & 27.17 & 36.50  \\ 
         & \name & \textbf{23.69} & \textbf{36.48} \\ 
        \midrule
        \multirow{2}{*}{\shortstack[l]{7 B}}  
        & GaLore & - & 22.9 \\
         & \name & - & \textbf{21.2} \\
         \bottomrule 
    \end{tabular}
\end{table}

We present the results in Table~\ref{tab:galore}. We notice that GaLore shows a similar level of memory usage as \name with a small overhead. This difference in memory is likely because GaLore stores additional projection matrices. Further, our \name achieves a better performance in terms of perplexity, suggesting its strong capability in optimization.

As discussed in Section~\ref{sec:related}, our \name is orthogonal to many memory-efficient automatic differentiation techniques. To show this, we additionally enable the activation checkpointing~\cite{chen2016training} and layer-by-layer updating like LOMO~\cite{lv2023full} for the 7B model. Specifically, activation checkpointing recomputes the activations during back-propagation instead of storing them in the forward pass. LOMO, on the other hand, promptly updates the layer weight upon obtaining its gradient without waiting until all layers are finished. Both techniques are unaffected when using \name. As a result, the peak memory allocated is only 21.2GiB for the 7B model, surpassing GaLore's memory saving of the same setting.

\subsection{Profiling results}~\label{apx:exp:prof}

\begin{figure}[h]
\centering
\begin{subfigure}{\textwidth}
    \centering
    \includegraphics[width=\textwidth]{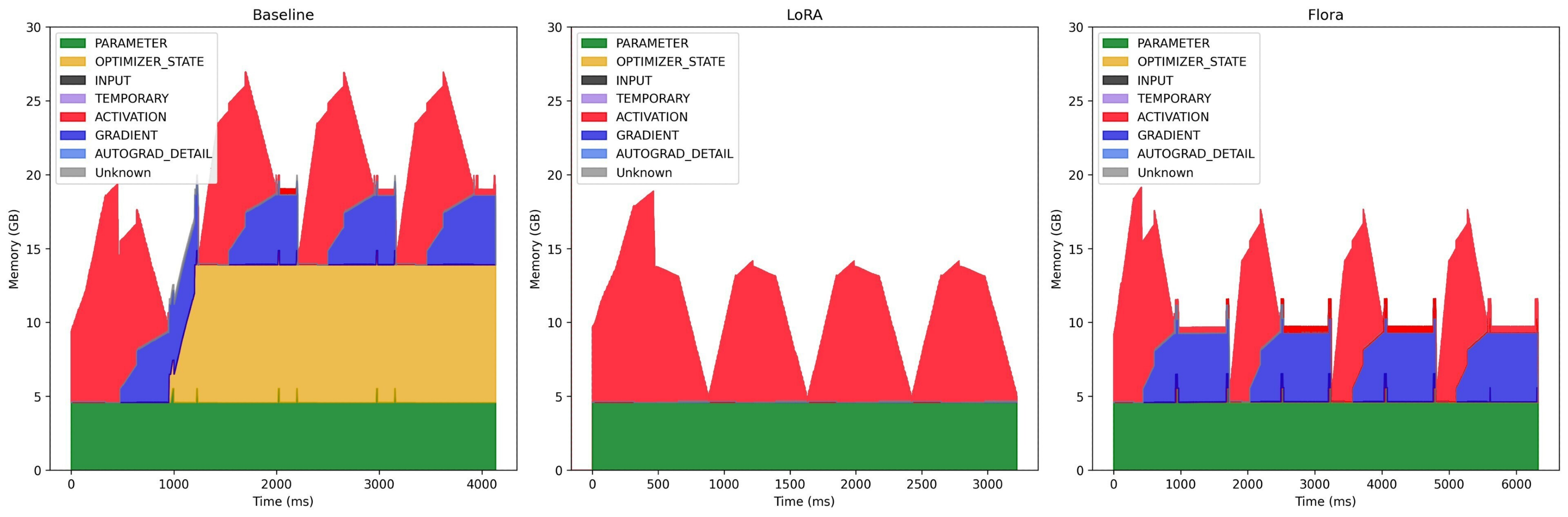}
    \subcaption{The vanilla training.}
    \label{fig:prof:full}
\end{subfigure}

\begin{subfigure}{\textwidth}
    \centering
    \includegraphics[width=\textwidth]{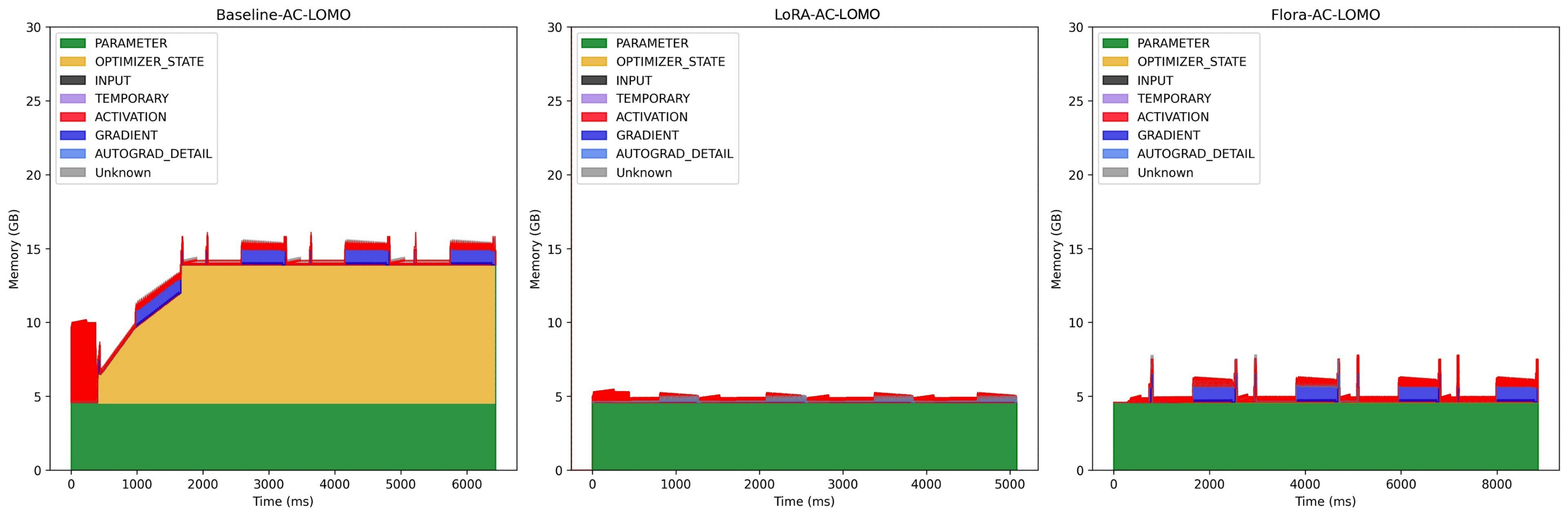}
    \subcaption{AC and LOMO enabled.}
    \label{fig:prof:lomo}
\end{subfigure}
\caption{Profiling the memory usage by categories during four iterations of training steps.}\label{fig:prof}
\end{figure}

We theoretically analyzed the memory usage of \name and LoRA in Section~\ref{sec:approach:flora}. In particular, we assert that both \name and LoRA reduce memory usage by maintaining smaller optimization states. To empirically verify this, we conduct the memory profiling analysis throughout the training time.
Specifically, we perform 4 training steps of a T5 model for all methods, including the vanilla training with Adam, LoRA, and \name. The sequence length is padded to 512, and the batch size is set as 4. We categorize the memory usage following PyTorch's convention. Additionally, we evaluate these methods in the setting where more memory-efficient training techniques like activation checkpointing~\cite{chen2016training} and LOMO~\cite{lv2023full} are enabled. We plot the profiling results in Figure~\ref{fig:prof}. 

In Figure~\ref{fig:prof:full}, we observe that both LoRA and \name have negligible memory footprint of the optimization states. Although \name uses a larger gradient storage, the peak memory is dominated by activations rather than gradients, resulting in a similar amount of peak memory usage. Further, as observed in Figure~\ref{fig:prof:lomo}, the difference in the gradient storage becomes less significant when activation checkpointing (AC) and LOMO are applied. In this case, both LoRA and \name demonstrate a similar memory usage pattern. We additionally provide the animation for the procedure for better illustration.\footnote{Please refer to our repository at \url{https://github.com/BorealisAI/flora-opt}.}

\end{document}